\documentclass[10pt,twocolumn,letterpaper]{article}

\usepackage{cvpr}              % To produce the CAMERA-READY version
\usepackage{graphicx}
\usepackage{amsmath}
\usepackage{amssymb}
\usepackage{booktabs}
\usepackage{float} 
\usepackage{wrapfig}

\usepackage[pagebackref,breaklinks,colorlinks]{hyperref}

\usepackage[capitalize]{cleveref}
\crefname{section}{Sec.}{Secs.}
\Crefname{section}{Section}{Sections}
\Crefname{table}{Table}{Tables}
\crefname{table}{Tab.}{Tabs.}

\def\cvprPaperID{*****} % *** Enter the CVPR Paper ID here
\def\confName{CVPR}
\def\confYear{2024}

\begin{document}

\title{Fill in the \underline{\ \ \ \ \ \ \ \ } (a diffusion-based image inpainting pipeline)}

% \author{Eyoel Gebre\\
% University of Washington\\
% {\tt\small eyoel23@cs.washington.edu}
% \and
% Krishna Saxena\\
% University of Washington\\
% {\tt\small ksaxena3@cs.washington.edu}
% \and
% Timothy Tran\\
% University of Washington\\
% {\tt\small timttran@cs.washington.edu}
% }

\author{Eyoel Gebre$^*$, Krishna Saxena$^*$, Timothy Tran$^*$\\
University of Washington\\
{\tt\small \{eyoel23, ksaxena3, timttran\}@cs.washington.edu}
}

\maketitle

\def\thefootnote{*}\footnotetext{These authors contributed equally to this work}\def\thefootnote{\arabic{footnote}}

%%%%%%%%% ABSTRACT
\begin{abstract}

Image inpainting is the process of taking an image and generating lost or intentionally occluded portions. Inpainting has countless applications including restoring previously damaged pictures, restoring the quality of images that have been degraded due to compression, and removing unwanted objects/text. Modern inpainting techniques have shown remarkable ability in generating sensible completions for images with mask occlusions. In our paper, an overview of the progress of inpainting techniques will be provided, along with identifying current leading approaches, focusing on their strengths and weaknesses. A critical gap in these existing models will be addressed, focusing on the ability to prompt and control what exactly is generated. We will additionally justify why we think this is the natural next progressive step that inpainting models must take, and provide multiple approaches to implementing this functionality. Finally, we will evaluate the results of our approaches by qualitatively checking whether they generate high-quality images that correctly inpaint regions with the objects that they are instructed to produce.
\end{abstract}

%%%%%%%%% BODY TEXT
\section{Introduction}
\label{sec:intro}

\subsection{Motivation}

Inpainting has been an important problem within the field of computer vision for many decades. It is a functionality that is essential to many image related applications such as object removal, image restoration, manipulation, re-targeting, compositing, and image-based rendering. It is also safe to believe that with the wide adoption of generative AI tools, functionalities like inpainting could see a significant increase in use within creative applications as well. As tools such as DALL-E, ChatGPT, Midjourney, and Sora are proving that they can be of high utility to those whose career depends on creative works, we are seeing for the first time how AI systems are significantly changing a large industry (The U.S. Bureau of Economic Analysis reports that arts and cultural production accounts for \$1,016,249,142,000 and 4.4\% of the U.S. economy, contributing 4,851,046 jobs ~\cite{NASAA2020}. Therefore, improving functionalities such as inpainting and allowing them to have a more robust range of capabilities can have a true economic impact. 

\subsection{Background}
There are two primary general approaches to the problem of inpainting: stochastic and deterministic. Stochastic methods output multiple sensible inpainting results through a random sampling process, while deterministic methods produce a single result. A common deterministic approach is to take an image along with a binary mask representing the region to be inpainted and pass these two images into a generator model such as a GAN trained to fill in the missing region. The task of improving the generator's ability to produce acceptable inpaintings has been tackled from a wide range of approaches including attention mechanisms, encoder-decoder connections, deep prior guidance, and multi-scale aggregation ~\cite{quan2024deep}. Stochastic inpainting techniques, such as flow-based and MLM-based methods, utilize generative models and sequence prediction to reconstruct image structures and textures. Additionally, many stochastic techniques take the approach of starting with a noisy image and iteratively denoise it until reaching a sensible output, a strategy commonly used by diffusion-based inpainting techniques which we will focus on.

\section{Related Work}
\label{sec:intro}

As mentioned, many deterministic strategies for implementing inpainting take an image dataset along with masks generated from some mask distribution and then train a model to fill in these masks. An example of this is Large Mask Inpainting (LaMa) ~\cite{LaMa}. During training, LaMa takes in sample images, and for each image generates a mask from some fixed mask distribution. Then, it trains a fast Fourier Convolutional Network to predict what is hidden behind the mask for each image, which is able to achieve strong results. However, the downside is that the model often struggles to generalize its abilities to masks which are not within the distribution used to generate masks in the training dataset. 

The current leading effort in inpainting that we have been able to identify is a paper from ETH Zurich called RePaint ~\cite{A2}. RePaint proposes a Markov Chain based random walk to add Gaussian noise to an input image followed by a reverse walk to denoise the masked portion of the image. This way, RePaint is able to remove the dependency on the specific mask distribution used to train the model by utilizing a pre-trained Denoising Diffusion Probabilistic Model (DDPM) to denoise pixels inside an arbitrary mask alongside their context (the unmasked portion of the original image with similar noise). Thus, RePaint gets around the lack of mask generalization all without even needing a training process by simply adjusting the input to an existing pre-trained DDPM. To add more semantic variance to the model’s output, the RePaint authors proposed a resampling step during the reversed random walk. In the resampling step, Gaussian noise is added back to a partially denoised combination of the learned masked region and noisy input unmasked region. This noisier combination is autoregressive and is passed into the DDPM a fixed number of times to extract the full benefit of the DDPM’s stochasticity, which is explained in further detail in Section \ref{subsec:ResamplingJumping}. 

In short, RePaint emphasizes denoising on the masked region while using the unmasked region mostly as a reference that provides context for the image. This results in very natural looking inpainted images which were evaluated higher than state-of-the-art models for a wide variety of mask distributions. Uniquely, RePaint is more creative with the objects it is able to inpaint while still allowing its final outputs to be semantically correct ~\cite{A2}. 

One unique feature of RePaint is that it often introduces random objects in its generated regions. Although these objects still often look natural and semantically correct, their randomness may be unwanted in many situations. For example, a lizard is generated when a person wants to remove a human hand from a portion of an image. We aim to allow RePaint to maintain its level of performance and semantic correctness while allowing for more control over what exactly gets generated, particularly in the masked region. RePaint currently does not take into consideration any preferences for what should be inpainted. There is also no clear interface that is best suited for providing the context/information to the model on what our preferred inpaint will contain.

Thus, our goal is to extend the abilities of RePaint such that in addition to taking in a mask and an image, it can also take in information regarding what is preferred to be inpainted. There are countless options: providing a textual description of what our target inpaint contains, an image of a preferred object to be generated, a dataset of acceptable objects that will allow the model to choose the most suitable option, etc. 

One benefit of RePaint is that it is only modifying the denoising step of an existing pre-trained diffusion model. This also means that we would not need to label any data or train the model, making it quite straightforward to try a large variety of these options for providing our preferred inpainted object and evaluating each approach. We will initially focus on evaluating the performance of RePaint when it is given just a single image containing what we want to be inpainted into the masked region. This will be done by providing the target image (i.e. a cutout of a dog) in the sampling (denoising) step of the RePaint algorithm similar to how the current algorithm provides a reference to the original unmasked region. In short, we will create a new pipeline where given the scene background, mask, and a third image representing a "target" object, RePaint will inpaint the occluded portion of the scene with an instance of the "target" object in a sensible manner. 

We will try various approaches to learning the borders of our target so that the model generates transitions from scene to target which are sensible. Note that one limitation of our model is that there is no ground truth to each image inpainting, making it difficult to quickly analyze the performance. We will instead evaluate the performance of this adjustment by manually comparing it to the output of the original RePaint model and possibly other leading inpainting models with the additional benchmark of whether it was able to generate our preferred object while maintaining an equal level of quality and semantic correctness. We expect this approach to work quite well as it does not make any assumptions that the original RePaint algorithm doesn’t as well.
\section{Method}
\begin{figure}[b]
    \centering
    \includegraphics[width=0.5\textwidth]{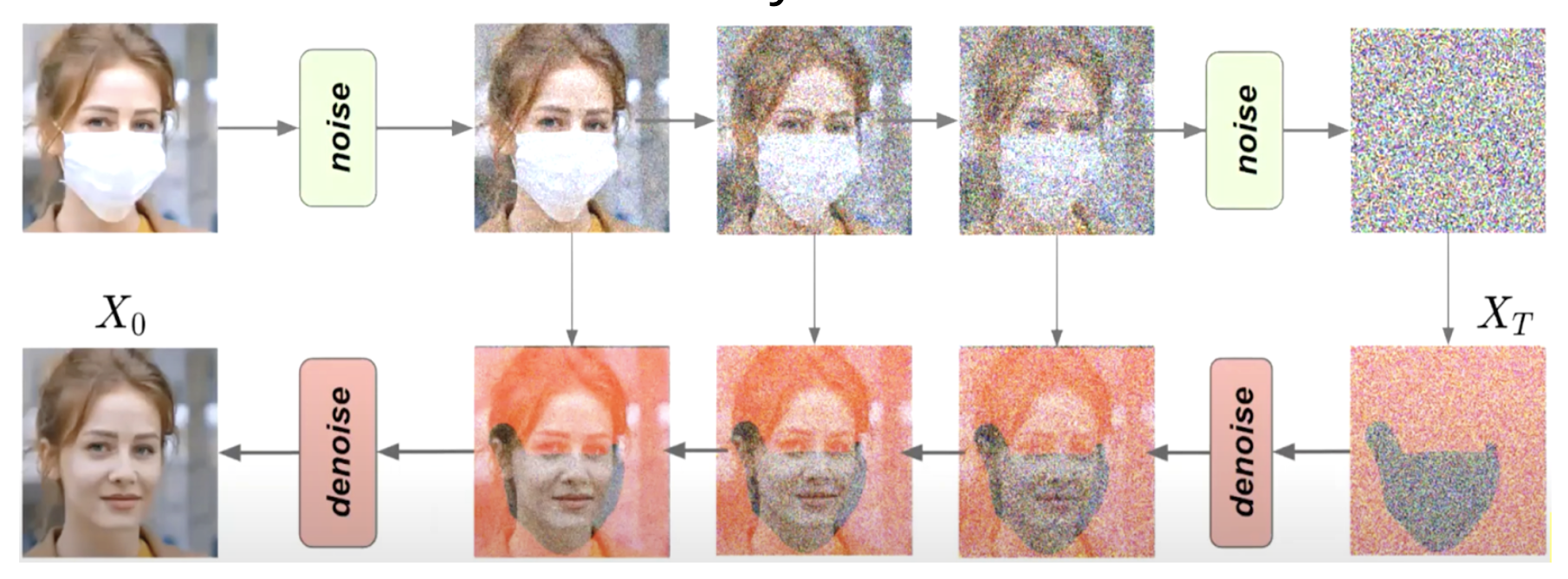}
    \caption{Pipeline diagram of RePaint (without sampling or jumping) reproduced from [Lugmayr et. al., 2022] for illustrative purposes.}
    \label{fig:repaintpipeline}
\end{figure}

\begin{figure*}[t]
    \centering
    \includegraphics[width=0.95\textwidth]{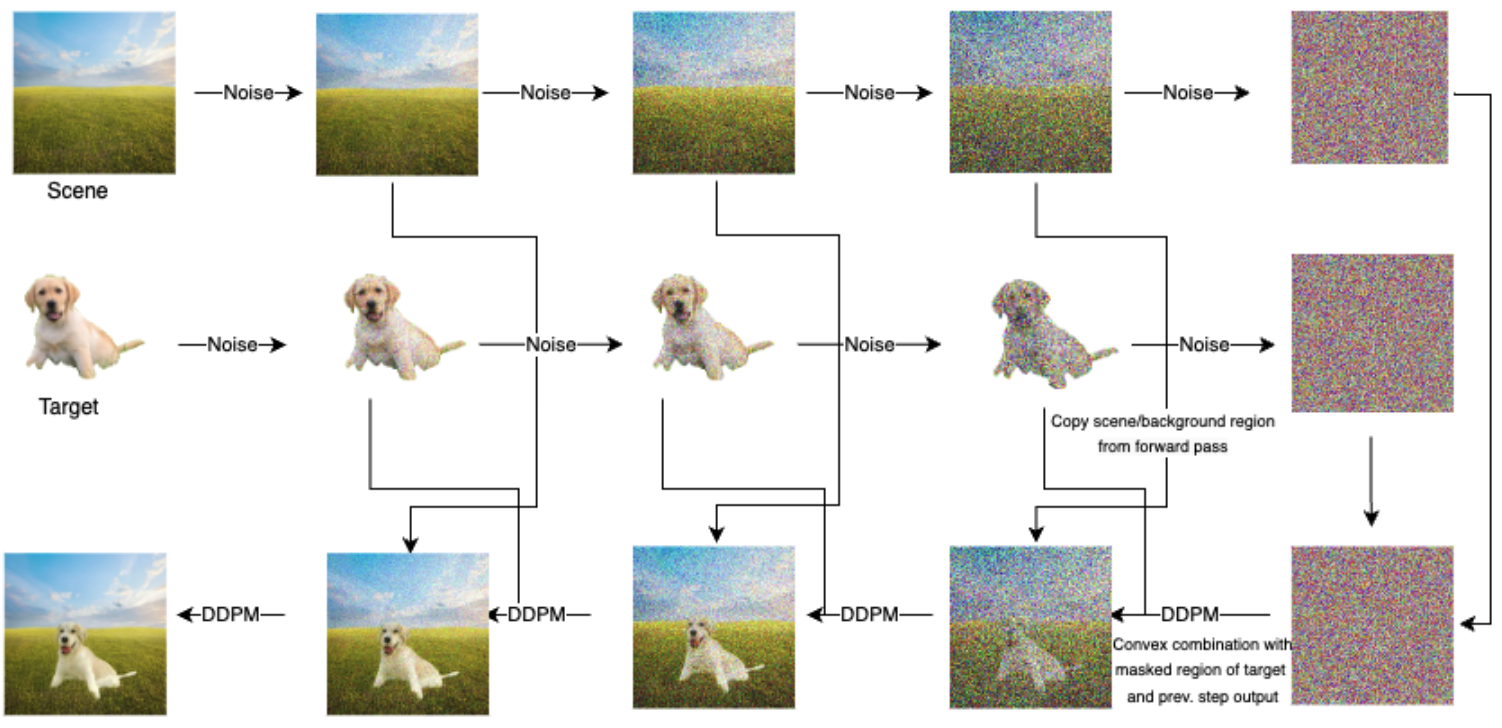}
    \caption{Core pipeline without resampling or jumping.}
    \label{fig:corepipeline}
\end{figure*}

Our pipeline builds on RePaint's inference algorithm (Section \ref{subsec:RepaintPipeline} and Figure \ref{fig:repaintpipeline}) by allowing the inpainting to be guided by a target image. To test our pipeline, we created a small dataset comprising of scene and target images alongside manually created binary masks for where the target will be inpainted on the scene. These three images (scene, target, and mask) become the inputs to our pipeline and get processed into one image as described in Section \ref{subsec:CorePipeline}.

\subsection{RePaint Pipeline Background}
\label{subsec:RepaintPipeline}
An in-depth overview of RePaint’s pipeline is necessary to understand our contribution. Initially, the diffusion process transforms the masked starting image $x_0$, into Gaussian white noise $x_T$ by adding iid Gaussian noise with progressively greater variance to a scene image. At each time step $t$ of the forward pass, the image undergoes transformations according to the diffusion process, becoming noisier as $t$ increases until it reaches pure noise at high $t$ values.

In the backward pass, the goal is to reconstruct the original image from the white Gaussian noise. At each time step $t$, the noisy image is inputted into a DDPM. The output of the masked-out portion of the image is extracted and combined with the non-masked region obtained from the corresponding forward pass step. This process gradually restores the original image by filling out the masked-out portion as accurately as possible.  (see Figure \ref{fig:repaintpipeline} for the pipeline diagram of RePaint). RePaint also introduces jumping and sampling to improve results, which will be explained in Section \ref{subsec:ResamplingJumping}.

\subsection{Core Pipeline for Inpainting Target Images}
\label{subsec:CorePipeline}
Our contribution is introducing the ability to specify an object to be inpainted into the scene at a specific location. The pipeline is designed to take in three input images: the background scenery, the target image, and a binary mask representing the region of interest within the target image. Notably, unlike RePaint’s pipeline, which solely processed the scene alongside an arbitrary mask, our approach incorporates contextual information for more target inpainting.
In addition to denoising the scene during the forward pass, our method also applies denoising to the target image using a similar procedure. During the backward pass, the image at timestep $t$, $x_t$, is passed to the DDPM as normal. This is described in equations 1-2, which were inspired by equations 8a-c in  \cite{A2}:
\begin{align}
    x_{\text{scene}, t-1} &\sim \mathcal{N}\left( \sqrt{\bar{\alpha}_t} x_{\text{scene}, 0}, \left(1- \bar{\alpha}\right)I\right) \tag{1a} &\label{eq:1a} \\
    x_{\text{target}, t-1} &\sim \mathcal{N}\left( \sqrt{\bar{\alpha}_t} x_{\text{target}, 0}, \left(1- \bar{\alpha}\right)I\right) \tag{1b} &\label{eq:1b}\\
    x_{\text{repaint}, t-1} &\sim \text{DDPM}\left( x_t \right) \tag{2} &\label{eq:ddpm}
\end{align}
where $\bar{\alpha}_t := \Pi_{i=1}^T (\beta_i)$. \\

While the unmasked region of the scene is extracted from the corresponding forward pass, the same as RePaint, a notable issue arises in handling the masked target, since we have both the masked noised object from the forward pass and the masked object from the DDPM. This discrepancy, which we call “mask conflict”, arises from the coexistence of two masked versions of the object: one generated during the forward pass and the other by the DDPM. To resolve this conflict and produce the image at $x_{t-1}$, we take a convex combination of the two masked target images using a parameter series $\lambda_t$, described as
\begin{align}
    x_{t-1}^{\text{unknown}} &= \lambda x_{\text{repaint}, t-1} + \left(1-\lambda \right)x_{\text{target}, t-1} \label{eq:maskconflict} &\tag{3}
\end{align}

Finally, to achieve our image $x_{t-1}$ in the backward step pass, we take the scene from the forward pass and the result of the "mask conflict" using the binary mask,
\begin{align}
     x_{t-1} &= 
    m \odot x_{\text{scene}, t-1} + \left(1-m \right) \odot x_{t-1}^{\text{unknown}} \label{eq:x_t-1} &\tag{4}
\end{align}
where $m$ represents the binary mask. This entire process is described in Figure \ref{fig:corepipeline}.

\subsection{Resampling and Jumping}
\label{subsec:ResamplingJumping}
In the pipeline (Figure \ref{fig:corepipeline}), each denoising (backward) step is a function of only the noised scene image, noised target image, and previous noised combination. During the linear combination to calculate the DDPM input, there exists the possibility that the border between the target area and masked, noised scene image has a sudden, unnatural change in color. The RePaint authors noted a similar issue with the combination of the previously noised combination and the masked scene and proposed resampling to address this issue. Resampling is a trick to increase the diversity and smoothness of the inpainted image by noising it and running it through the DDPM multiple times. This allows the DDPM to predict the pixels on both sides of the mask's boundary which increases the quality of these regions (along with the added benefit of increasing the variance of pixel values generated inside the mask). For example, first, the linear combination of a noisy, masked scene and generated image is passed into the DDPM. The result of this operation is a slightly less noisy combination with jarring boundaries. Gaussian noise is added to this result which is then passed back into the DDPM. This process is repeated $r$ times, after which the boundary transition is smoother. Since running the DDPM is time and computationally-intensive, the RePaint authors propose a jumping schedule parameter $j$ to control when resampling occurs. Every $j$ timesteps of the backward pass, $r$ resampling steps occur. When $r<=j$, the pipeline has the same runtime and FLOPs as the original pipeline ($r,j=1$).

\subsection{Initial Hyperparameter search}
\label{subsec:keyHparams}
As discussed in Section \ref{subsec:CorePipeline}, $\lambda_t$ controls the convex combination of the generated inpainted target with the noised ground truth target at timestep $t$. Note that with $\lambda_t = 1$, our pipeline collapses into RePaint’s original pipeline since there would be no contribution from the noised target in the forward step target images. With this pipeline design, we had an initial hyperparameter search over various values for $\lambda_t \in [0.8, 0.9, 0.993, 0.995, 0.999, 0.9999]$. Additionally, for the jump length and jump size variables, we tried values in the range $[10, 20, 30, 40]$. Finally, for the timestep count, we tried values in the range $[50, 100, 150, 200, 250]$. 

As expected our result ends up looking more like random noise or takes on an uninterpretable shape the closer our lambda value is to one. However, smaller lambda values result in an output that looks like the target image was simply cropped out and pasted on top of the scene background. We found that lambda values in the range of 0.92-0.97 performed the best at keeping the content of the target image without making it seem like it was an exact copy and paste.
Additionally, we expected to see that as our sampling parameter ($t$) increases to a large value, our model’s output is less fuzzy/noisy. This makes sense since this would put less weight on our forward pass noised target images in the denoising steps.

Finally, we noticed that increasing our jump and jump length parameters allowed for all-around higher quality outputs. This makes sense since the whole purpose of jumping is to allow the model to generate output which takes into account the surrounding scene’s context as much as possible when having the DDPM generate output. In summary, we performed a standard grid search over the ranges of all these parameters and identified the combination of optimal values to be 40 for both jump length and jump size, 200 for timesteps, and 0.993 for $\lambda_t$.

\newlength{\imagewidth}
\setlength{\imagewidth}{0.15\textwidth}

\begin{figure}
  \centering
  % First row of images
  \begin{subfigure}[b]{\imagewidth}
    \includegraphics[width=\textwidth]{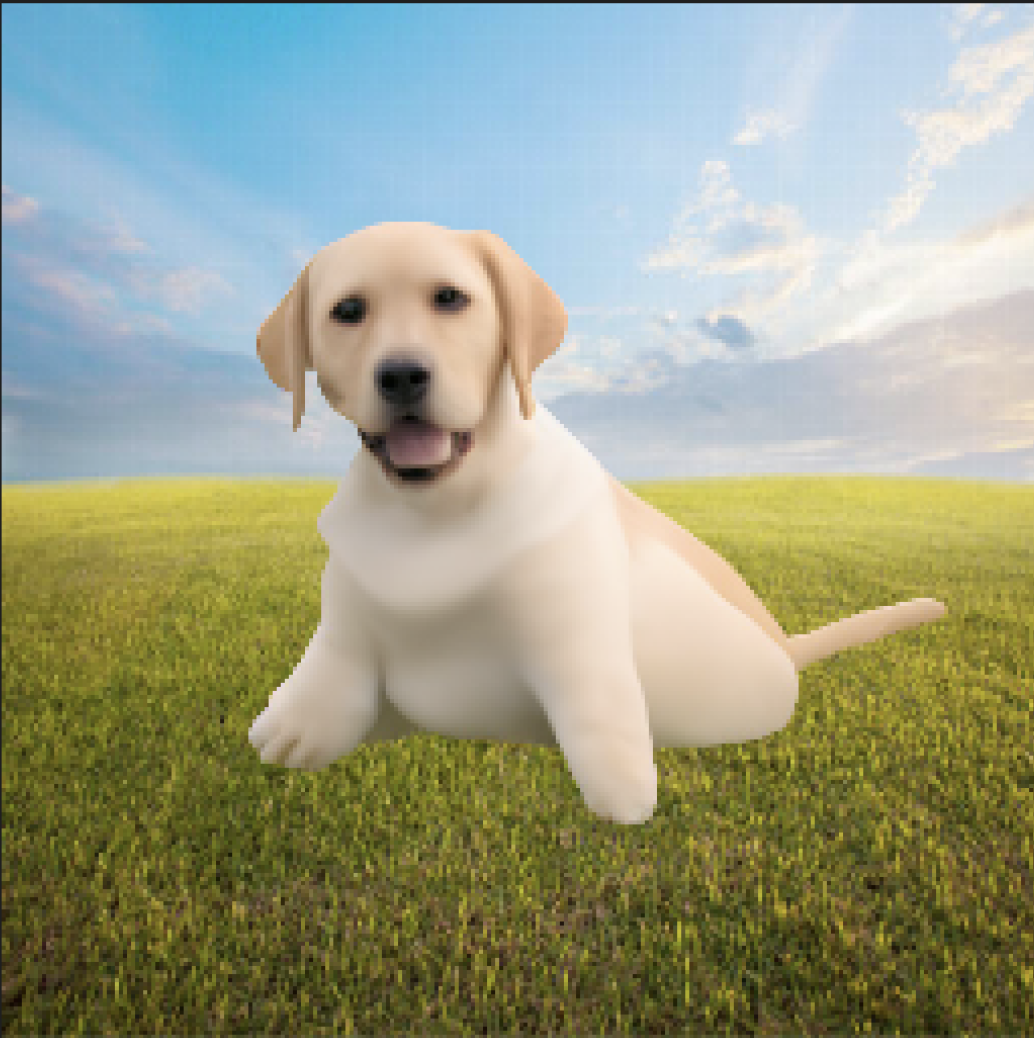}
    \caption{.8, 250, 20}
    \label{fig:image1}
  \end{subfigure}
  \hfill % adds horizontal space between the images
  \begin{subfigure}[b]{\imagewidth}
    \includegraphics[width=\textwidth]{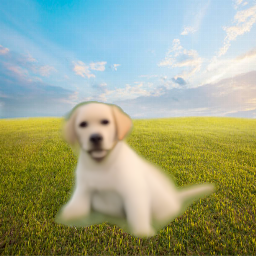}
    \caption{.993, 100, 30}
    \label{fig:image2}
  \end{subfigure}
  \begin{subfigure}[b]{\imagewidth}
    \includegraphics[width=\textwidth]{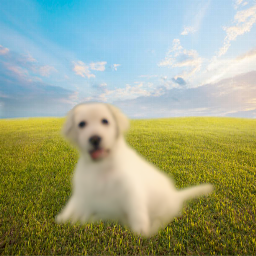}
    \caption{.997, 100, 40}
    \label{fig:image3}
  \end{subfigure}
  
  \hfill % adds horizontal space between the images
  \begin{subfigure}[b]{\imagewidth}
    \includegraphics[width=\textwidth]{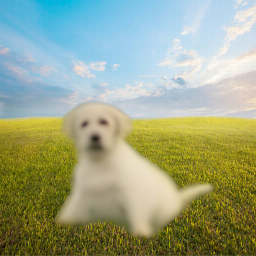}
    \caption{.999, 250, 40}
    \label{fig:image4}
  \end{subfigure}
  \hfill % adds horizontal space between the images
  \begin{subfigure}[b]{\imagewidth}
    \includegraphics[width=\textwidth]{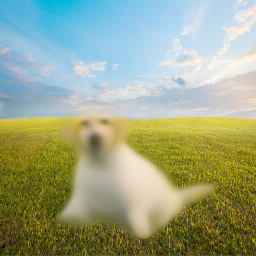}
    \caption{.999, 100, 40}
    \label{fig:image2}
  \end{subfigure}
  \begin{subfigure}[b]{\imagewidth}
    \includegraphics[width=\textwidth]{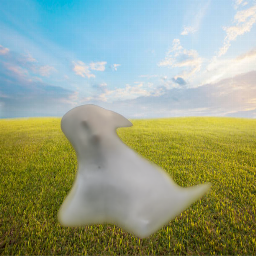}
    \caption{.80, 250, 20}
    \label{fig:image3}
  \end{subfigure}
  
  \caption{Hyperparameter search ($\lambda$, $t$, $j$), where $t$ is the number of noising timesteps in the forward pass and $j$ is both the number of jumps and jump length.}
  \label{fig:images}
\end{figure}

During our experimentation, we identified two viable methods for defining the target mask. The first approach involved creating an exact cutout of the target object, allowing the model to generate a more detailed image. However, this method often resulted in an unnatural-looking transition from the scene to the target object, appearing as if the object was simply pasted onto the image. Alternatively, we explored a more lenient approach by including additional information surrounding the target object in the mask. While this led to a more natural-looking transition between the scene and the target object, the resulting image of the target tended to be less detailed and smooth. Another issue with this approach was the dependency on the target image being in a similar environment to the scene. For example, if the target was a dog standing on gravel and the scene was a grass field, it would not make sense to give the model the grey-colored gravel border to inpaint into the final image. To address this, we investigated alternative modifications to the exact mask approach, as outlined in Section \ref{subsec:MaskAlt}.

\section{Experiments}
\label{sec:Experiments}
All though our initial results did not look like they were exact copy-pastes, we were still disappointed with the lack of interaction that our target object had with the background scene. An ideally inpainted object should interact with elements of the background scene at least to some degree. For example, in the case of the dog on the field, we would like to see the dog's paws and tail be hidden by blades of grass. However, our core pipeline cannot currently achieve this. We propose several modifications to our original pipeline to combat this issue.

\subsection{Masking Alternatives}
\label{subsec:MaskAlt}
The biggest issue we found from our results was the unnatural-looking boundaries in our unpainted images. Currently, with an exact binary mask, the pipeline didn't have context about how "close" a pixel was to the boundary. Thus, to help provide context and more flexibility to the DDPM in the boundary error, we decided to explore some alternative masking methods.

\subsubsection{Distance-Based "Heated" Mask}
To enhance boundary realism while still keeping the core target image, we adjusted the backward pass to use a "heated" buffer. Instead of using a binary buffer where we would take the value of every pixel of the target, the values of the mask were between 0 and 1. With this heated mask, we kept the forward pass of the target for pixels that were far away from the boundary (represented by values closer to 1) and allowed the DDPM the freedom and flexibility to generate natural-looking borders for the pixels of the target image close to the border.

For some integer $b$, the buffer size, the heated mask is a function of mask.
\begin{align}
     m_{i,j}^{\text{HEATED}} = \max{\frac{d_{i,j}}{b}, 1}  \label{eq:heated} &\tag{5}
\end{align}
% $$mask_{i,j}^{\text{HEATED}} = \max{\frac{d_{i,j}}{b}, 1}$$
where $d_{i,j}$ is the Manhattan distance from coordinate $(i,j)$ to the nearest black ($0$) pixel in the binary mask.

\subsubsection{Scene Buffer}
Secondly, to further improve boundary realism, we implemented a “scene buffer” technique by adding a 4 pixel border around the target mask during the linear combination. Here, an exact cutout of the mask is utilized, and during the backward pass, the linear combination is computed to resolve the mask conflict as previously described in \ref{eq:maskconflict}. The key distinction lies in incorporating a small border around the target from the DDPM output into the next backward step, instead of taking from the forward pass scene. Note that for the rest of the scene outside the small buffer, we take the forward pass from the corresponding timestep of the scene as normal. This modification removes the dependency for the target image to be in the same environment as the scene, while still allowing the model to create a natural looking transition at the border when using an exact cutout. This process is explained in the following equations, 
\begin{align}
    m_\text{ring} &= m_\text{ext} - m \tag{6}  \\
    x^\text{unknown, buf}_{t-1} &= (cx_{\text{repaint},t-1} + (1-c)x_{\text{target},t-1}) \tag{7} \\
    x_{\text{scene, buf},t-1} &= m_{\text{ext}} \odot x_{\text{scene},t-1} + m_\text{ring} \odot x^\text{unknown}_{t-1} \tag{8} \\
    x_{t-1}&= m \odot x_{\text{scene, buf}, t-1} (1 - m) \odot x^\text{unknown, buf}_{t-1} \tag{9}
\end{align}
where $m_\text{ext}$ represents the extended mask and $c$ is a new constant to replace $\lambda$, which is still used by $x^\text{unknown}_{t-1}$ from \ref{eq:x_t-1} for the buffer/ring area.

\subsection{$\lambda$ Scheduling to Increase RePaint Integration}
\label{subsec:lambda}
\begin{figure}[H]
  \centering
  \includegraphics[width=0.25\textwidth]{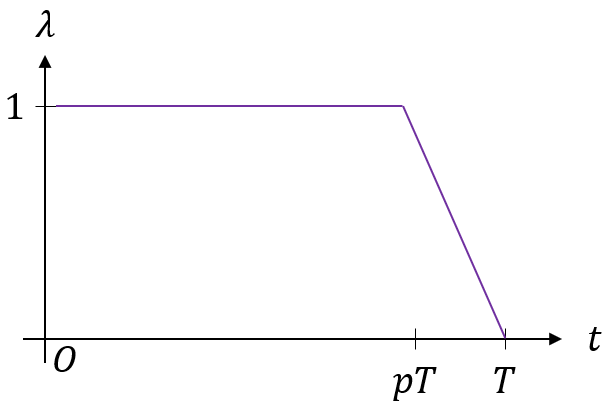}
  \caption{Schedule for $\lambda_t$ vs $t$}
  \label{fig:lambdaSched}
\end{figure}

Additionaly, we explored increased integration with RePaint’s model to address issues of lacking detailed features and producing smooth images with precise masks or jarring boundaries. This involves relying less on the noised target from the forward pass and more on the previous DDPM generation. Specifically, for timesteps $t$ closer to 0 in \ref{eq:maskconflict}, we set $\lambda_t=1$ so that the pipeline can infer and create a more natural border around the inpainted target portion for the remainder of the timesteps. While retaining the same input structure (scenery, target, mask target), we modified the backward pass to transition to RePaint’s model by scheduling $\lambda_t$ to be a linear interpolation \ref{fig:lambdaSched} from 0 to 1 for timesteps $T$ to $pT$, such that $p \in\left[0,1\right]$ and 1 for all $t<pT$.

Based on our results from Section \ref{subsec:keyHparams} (particularly note \ref{fig:images}), we let the hyperparameters $r,j=40$, reduce $T$ to $100$, solely to generate samples quicker for experimentation and conduct a grid search for $p \in { 0.1, 0.25, 0.5, 0.75, 0.9 }$.

We found that as $p$ increases, the inpainted image becomes less faithful to the target and more blurry. This is because a high value of $p$ implies that the pipeline is denoising without any direct contribution from the target image for $pT$ denoising steps.
For such denoising steps, the pipeline is equivalent to RePaint, so the generated images will have high variance as noted in Guided Diffusion \cite{dhariwal2021diffusion} and RePaint \cite{A2}.

\begin{figure}[H]
  \begin{subfigure}[a]{\imagewidth}
    \includegraphics[width=\textwidth]{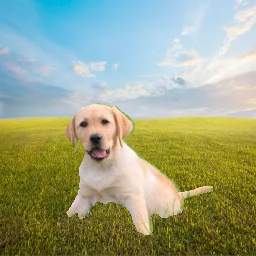}
    \caption{$p=0.1$}
    \label{fig:pExpImga}
  \end{subfigure}
  \hfill % adds horizontal space between the images
    \begin{subfigure}[a]{\imagewidth}
    \includegraphics[width=\textwidth]{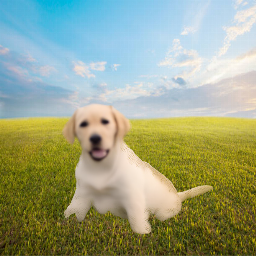}
    \caption{$p=0.25$}
    \label{fig:pExpImgb}
  \end{subfigure}
  \hfill
  \begin{subfigure}[a]{\imagewidth}
    \includegraphics[width=\textwidth]{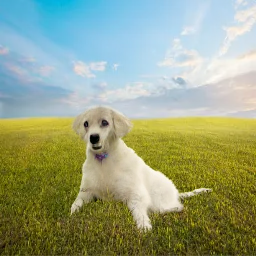}
    \caption{$p=0.5$}
    \label{fig:pExpImgc}
  \end{subfigure}
  \hfill
  \begin{subfigure}[a]{\imagewidth}
    \includegraphics[width=\textwidth]{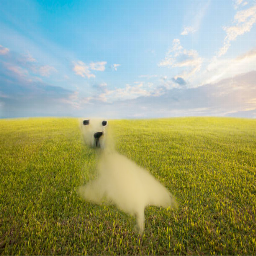}
    \caption{$p=0.75$}
    \label{fig:pExpImgd}
  \end{subfigure}
  \hfill
  \begin{subfigure}[a]{\imagewidth}
    \includegraphics[width=\textwidth]{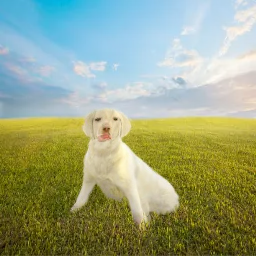}
    \caption{$p=0.9$}
    \label{fig:pExpImge}
  \end{subfigure}
  \begin{subfigure}[a]{\imagewidth}
    \includegraphics[width=\textwidth]{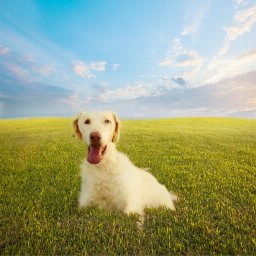}
    \caption{$p=0.5$ (2)}
    \label{fig:pExpImge}
  \end{subfigure}
  \caption{Images generated with $\lambda$ schedule for varying values of $p$, $T=100$, and $r, j=40$.}
  \label{fig:images}
\end{figure}

We identify the best value for hyperparameter $p$ as $0.5$ for the inpainting task. This setting allows for the inpainted image to be similar to the target image while allowing the pipeline to seamlessly fill in the outer border of the target subject. Additionally, this value allows the pipeline to retain the semantic meaning from the target image while inpainting novel and semantically consistent additions (e.g., the bow added to the dog in \ref{fig:pExpImgc} and the grass interacting with its paws)

\begin{figure*}
    \centering
    \includegraphics[width=0.95\textwidth]{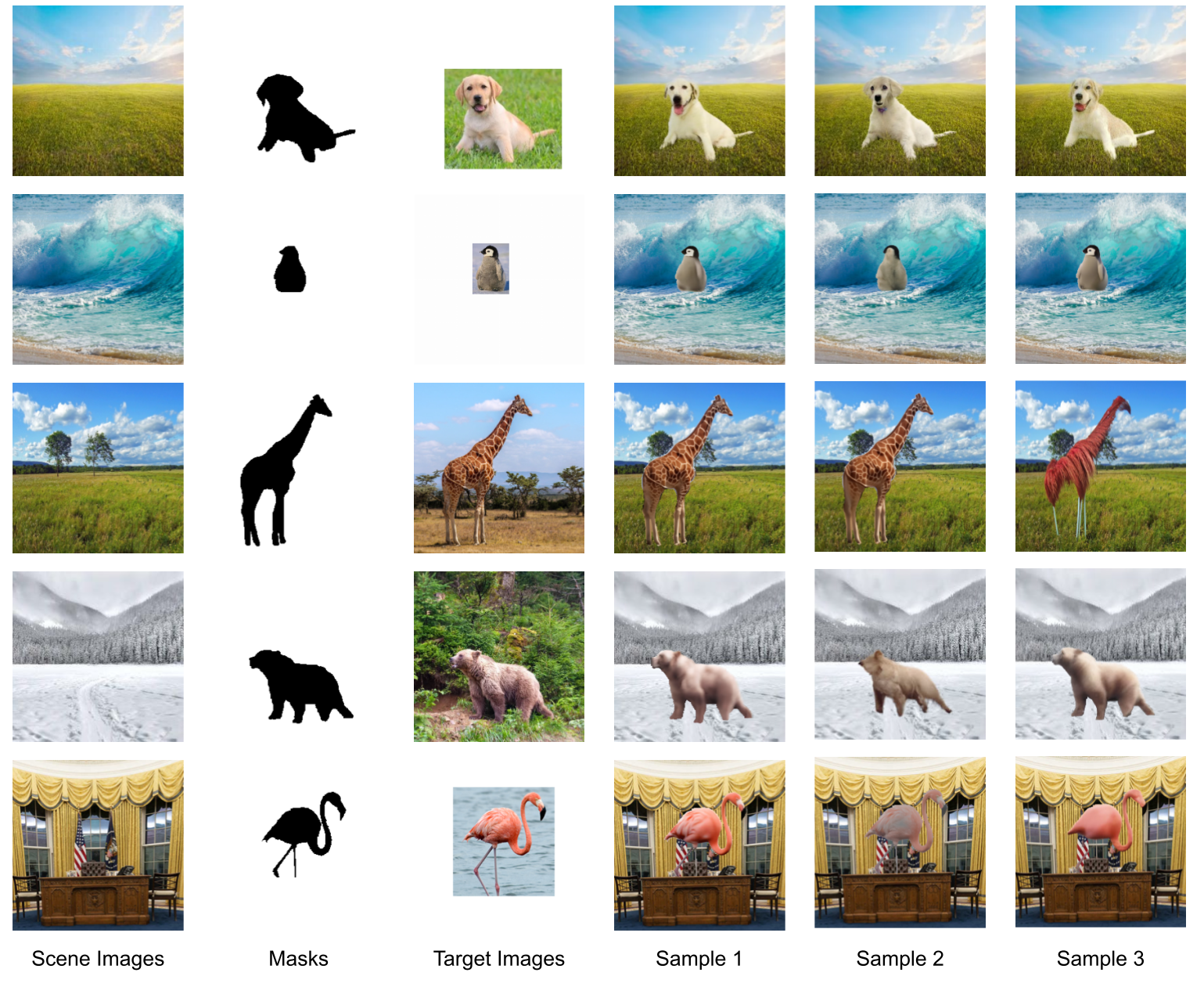}
    \caption{Example scenes, masks, and targets along with several sample outputs.}
    \label{fig:MainResults}
\end{figure*}

\subsection{Failure Modes}
As described previously, there is a high variance in the final inpainted when running with $p > 0.5$. The case when $p=0.5$ is particularly interesting because much of the semantic information of the target image is kept but the high number of denoising passes through the DDPM in this scenario makes the generations creative.

For a run with $T=100, r,j=40,$ and $p=0.5$, a giraffe target image and grassland scene (middle right image in Figure \ref{fig:MainResults}), the resulting image maintains the shape and color of a giraffe due to the masking $\forall t$ denoising steps but the content inside of the mask resembles a flamingo-giraffe as if it came from a Truffula forest scene from Dr. Seuss' \textit{The Lorax}. After rerunning with these hyperparameters multiple times, we could not produce a result that was so distinct from a giraffe. To mitigate this failure mode in a production setting, we recommend producing at least two candidate images per scene and target.

Another failure mode when $p>0.5$ is that this setting reveals the biases of the DDPM train data. As the RePaint authors note, a DDPM model trained on ImageNet, as is the one used to generate all images in this report, will be biased towards denoising dogs \cite{A2}. For iterations $t$ of our pipeline such that $\forall \tau < t, \lambda_\tau = 1$, the autoregressive nature of the denoising process means that all future steps in the denoising process are biased towards high likelihood images in the DDPM dataset.

To address this failure mode, we recommend using a model like ResNet \cite{he2015deep} to find the class of the target image and using a DDPM trained on a dataset where that image class is not underrepresented.

\subsection{Final Pipeline}

The authors present Fill in the \underline{\ \ \ \ \ }, a diffusion-based inpainting pipeline that seamlessly inserts target images into scenes. Based on the current analyses, the authors recommend the following hyperparameters: diffusion timesteps $T=200$, jumping steps $j=40$, resampling steps $r=40$, and a piecewise linear $\lambda$ schedule with $p=0.5$ (i.e., $\lambda_t = 1$ if $t \leq 0.5T$ and a linear interpolation from $(pT, 1.0)$ to $(T, 0)$ elsewhere.

\section{Conclusion and Acknowledgements}
\subsection{Further Steps}

Moving forward, there are several avenues for further exploration and improvement in our research. First, we can delve deeper into mask modifications, as described in Section \ref{subsec:MaskAlt}. The majority of our experimentation was with hand-made exact binary masks, but exploring alternative masking techniques such as gradient-based or buffered masks may allow the model to produce more realistic-looking inpaintings while still keeping the details of the original target. Furthermore, while the introduction of lambda scheduling in Section \ref{subsec:lambda} improved the performance of our core pipeline, the addition of a more dynamic-based lambda scheduling could further enhance the adaptability and versatility of the pipeline. This dynamic approach would involve adjusting the lambda values based on the characteristics of different scene-target pairs, therefore optimizing the inpainting process for each specific scenario.

In addition to exploring mask modifications and refining lambda scheduling, expanding our testing to encompass a wider variety of images would be a huge improvement. Currently, the manual creation of the target mask poses scalability challenges. A solution could be to leverage segmentation techniques that would automate mask generation, facilitating the creation and efficient testing on a larger dataset. With a larger dataset, it would be much easier to analyze a diverse range of scene-target pairs, leading to optimal hyperparameters that generalize well across various scenarios.

Finally, our research aims to achieve a fully automated pipeline for inpainting. This system would require minimal user input, consisting mainly of providing a scene and target images and selecting the inpainted region while automating the target position, mask creation, and choice of hyperparameters.

\subsection{Conclusion}
In this study, we aimed to improve image inpainting techniques by enhancing control over what exactly gets generated. In particular, we identified that current approaches did not allow a model to base its inpainted output on an image of a target object as opposed to textual prompting or other ways of specifying a desired inpainted object. Leveraging recent advancements in generative AI and inpainting methodologies, we focused on a diffusion-based approach where we modify the inputs given to a diffusion model on its denoising steps to feed it the target image's context. Through some adjustments and refinements, we sought to mitigate challenges such as mask conflicts and boundary realism, achieving some interesting results in the quality and naturalness of inpainted images with target objects. Our experiments provided valuable insights into the effectiveness of our approach and the importance of various hyperparameters. Overall, our project represents a modest yet meaningful step forward in the field of image inpainting.

\subsection{Acknowledgements}
We would like to kindly thank the CSE 493G staff for the amazing lecture content and guidance that allowed us to complete this project. Additionally, we express our appreciation to the RePaint paper and its authors, which was a huge inspiration to our paper.

%%%%%%%%% REFERENCES
{\small
\bibliographystyle{ieee_fullname}
\bibliography{egbib}

\begin{thebibliography}{1}\itemsep=-1pt

\bibitem{dhariwal2021diffusion}
Prafulla Dhariwal and Alex Nichol.
\newblock Diffusion models beat gans on image synthesis, 2021.

\bibitem{he2015deep}
Kaiming He, Xiangyu Zhang, Shaoqing Ren, and Jian Sun.
\newblock Deep residual learning for image recognition, 2015.

\bibitem{A2}
Andreas Lugmayr, Martin Danelljan, Andres Romero, Fisher Yu, Radu Timofte, and Luc~Van Gool.
\newblock Repaint: Inpainting using denoising diffusion probabilistic models, 2022.

\bibitem{NASAA2020}
{National Assembly of State Arts Agencies}.
\newblock Creative economy state profiles.
\newblock \url{https://www.nasaa-arts.org}, 2020.
\newblock Accessed: 2024-02-29.

\bibitem{quan2024deep}
Weize Quan, Jiaxi Chen, Yanli Liu, Dong-Ming Yan, and Peter Wonka.
\newblock Deep learning-based image and video inpainting: A survey.
\newblock Received: date; Accepted: date.

\bibitem{LaMa}
Roman Suvorov, Elizaveta Logacheva, Anton Mashikhin, Anastasia Remizova, Arsenii Ashukha, Aleksei Silvestrov, Naejin Kong, Harshith Goka, Kiwoong Park, and Victor Lempitsky.
\newblock Resolution-robust large mask inpainting with fourier convolutions, 2021.

\end{thebibliography}
}

\end{document}